\documentclass[twocolumn,10pt]{IEEEtran}
\usepackage{geometry}
\newgeometry{left=0.80in,right=0.80in,top=1.0in,bottom=1.0in}
\usepackage{graphics} 
\usepackage{amsmath} 
\usepackage{amsfonts}
\usepackage{bm}
\usepackage[caption=false]{subfig}
\usepackage{tikz}
\usepackage{pgfplots}
\DeclareMathOperator*{\argmax}{arg\,max}
\DeclareMathOperator*{\argmin}{arg\,min}
\usepackage{caption} 
\usepackage{lipsum}
\usepackage{cuted}
\usepackage{enumerate}
\usepackage{booktabs} 
\usepackage{graphicx}
\usepackage{float}
\usepackage{mathrsfs}

\usepackage{algpseudocode}
\usepackage{algorithm}
\algnewcommand\algorithmicforeach{\textbf{for each}}
\algdef{S}[FOR]{ForEach}[1]{\algorithmicforeach\ #1\ \algorithmicdo}
\usepackage{eqparbox}
\newdimen{\algindent}
\algnewcommand\LeftComment[2]{%
\hspace{#1\algindent}$\triangleright$ \eqparbox{COMMENT}{#2} \hfill %
}
\algnewcommand\LeftCommentNoTriangle[2]{%
\hspace{#1\algindent} \eqparbox{COMMENT}{#2} \hfill %
}

\begin{document}
\title{\LARGE \bf
On the Robustness of Deep Reinforcement Learning in  IRS-Aided Wireless Communications Systems 
}
\author{\IEEEauthorblockN{Amal Feriani, Amine Mezghani, \textit{Member, IEEE}, and Ekram Hossain, \textit{Fellow, IEEE}}\thanks{The authors are with the Department of Electrical and Computer Engineering at the University of Manitoba, Canada (emails: \{feriania@myumanitoba.ca, Amine.Mezghani@umanitoba.ca, Ekram.Hossain@umanitoba.ca \} }}

\maketitle

\begin{abstract}

We consider an Intelligent Reflecting Surface (IRS)-aided multiple-input single-output (MISO) system for downlink transmission. We compare the performance of Deep Reinforcement Learning (DRL) and conventional optimization methods in finding optimal phase shifts of the IRS elements to maximize the user signal-to-noise (SNR) ratio. Furthermore, we evaluate the robustness of these methods to channel impairments and changes in the system. We demonstrate numerically that DRL solutions show more robustness to noisy channels and user mobility.
\end{abstract}


\section{Introduction}
Recently, Deep Reinforcement Learning (DRL) has attracted a lot of interest to solve a wide range of wireless communications problems~\cite{9174775,9324737}. Thanks to its learning capabilities and inference speed, DRL has shown competitive performance compared to traditional optimization methods. This is why DRL-based algorithms are considered as one of the key technologies for future wireless generations. In the same vein, Intelligent Reflecting Surfaces (IRS) represent a promising pillar of the future wireless evolution since they are able to provide additional gain in wireless system performance at a low cost. 

DRL has been already applied to solve IRS-related problems such as passive/active beamforming. In \cite{feng2020deep}, the authors considered a Multiple-Input Single-Output (MISO) system and used DRL to find optimal passive phase shifts. The DRL agent achieves a better performance than the fixed-point iteration algorithm and  results close to a SemiDefinite Relaxation (SDR) upper bound. The joint optimization of transmit beamforming and phase shifts in multi-user MISO network was proposed in \cite{huang2020MultiUserMiso} where the authors used a unified DRL algorithm to solve it. Furthermore, \cite{lin2020IRSdrl} approached the problem of minimization of the Base Station (BS) transmit power in MISO system using DRL for passive beamforming and traditional optimization-based method for active beamforming. An IRS-assisted downlink NOMA multi-user system was studied in \cite{shehab2021DRLNOMAIRS} and DRL was employed to predict the IRS phase shifters. DRL showcases better sum-rate performance compared to the OMA scheme.

The objective of this paper is twofold: First, we provide a comparison between DRL methods and the best performing traditional optimization methods such as Vector Approximate Message Passing (VAMP) \cite{rangan2019VAMP} and Alternating Direction Method of Multipliers (ADMM) \cite{boyd2011ADMM} in solving the phase shift optimization problem for IRS-aided wireless communications systems. This work will help to build a benchmark to compare DRL to other exiting techniques. Furthermore, unlike the works mentioned above, our aim is not to solve passive beamforming with DRL but to motivate the use of DRL over other conventional methods in practical settings. For instance, we focus on the robustness of the optimal solution to channel impairments. To do so, we compare the decrease in the achieved SNR when the Channel State Information (CSI) is perturbed (i.e. additional noise) and the user location changes. Our work is novel since we consider that the agent's knowledge of the environment is not perfect and we test the obtained solution under different perturbation scenarios. Our objective through this work is to bring a more practical perspective to the proposed algorithms for phase shift optimization and to wireless problems in general. Simulation results show that, compared to other optimization methods, DRL is more robust  to changes in the system. This is another advantage of DRL methods along with the inference speed that motivates the use of learning-based techniques to solve wireless problems.

The rest of the paper is organized as follows. The  system model, assumptions, and the problem formulation are presented in Section~II. The DRL-based approach for phase-shift optimization is
presented in Section III. The performance evaluation and benchmarking results are presented in Section~IV before the paper is concluded in Section~V.

\section{System Model and Problem formulation}
\begin{figure}[ht!]
    \centering
    \includegraphics[scale=0.5]{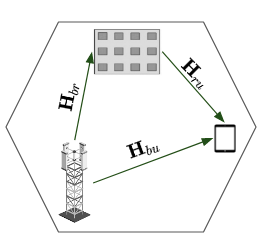}
    \caption{IRS-aided single-user MISO system.}
    \label{fig:miso}
\end{figure}

Consider a BS equipped with $N$ antenna elements serving a single-antenna user in the downlink. An IRS, with $M=M_x \times M_y$ reflective elements, is deployed to assist the BS. $M_x$ and $M_y$ refer to the number of IRS elements in each row and column, respectively. We denote by $\mathbf{H}_{bu} \in \mathbb{C}^{N\times 1}$, $\mathbf{H}_{ru} \in \mathbb{C}^{M\times 1}$, and $\mathbf{H}_{br} \in \mathbb{C}^{M\times N}$ the direct channel between the BS and the user, the channel between the IRS and the user and the MIMO channel linking the BS and the IRS (Fig.~\ref{fig:miso}).
\noindent Henceforth, the received signal at the user is given by:
\begin{align*}
    y = (\mathbf{H}_{bu}^{H} + \mathbf{H}_{ru}^{H}\mathbf{\Phi} \mathbf{H}_{br})\mathbf{F}s + n
\end{align*}
where $s$ is a zero-mean and unit-variance transmit symbol, $n \sim \mathcal{CN}(0, \sigma^2_n)$ is the additive white Gaussian noise. $\mathbf{F} \in \mathbb{C}^{N\times 1}$ is the beamforming vector satisfying the constraint $||\mathbf{F}||^2=P_{max}$ with $P_{max}$ being the BS maximum transmit power. $\mathbf{\Phi} = Diag(\phi_1, \dots, \phi_M) \in \mathbb{C}^{M\times M}$ is the IRS phase shift matrix such that $\phi_i = e^{j\theta_i}$ with $\theta_i \in [0, 2\pi]$ representing the phase shift of the $i$th IRS reflective element for $i=1, \dots, M$. 
Consequently, the user's effective channel is expressed as follows:
\begin{align*}
    \mathbf{H}^H = \mathbf{H}_{bu}^{H} + \mathbf{H}_{ru}^{H}\mathbf{\Phi} \mathbf{H}_{br}.
\end{align*}
Furthermore, we assume that the channels are quasi-static frequency flat-fading and the CSI is available at the BS.
The received SNR at the user is given by 
\begin{align}
    \rho = \frac{|\mathbf{H}^H\mathbf{F}|^2}{\sigma_n^2}. \label{eq:snr}
\end{align}

We seek to optimize the phase shift matrix $\mathbf{\Phi}$ and the beamforming vector to maximize the user SNR. The optimization problem can be stated as follows:
\begin{align*}
    \max_{\mathbf{\Phi}, \mathbf{F}} & \quad ||(\mathbf{H}_{bu}^{H} + \mathbf{H}_{ru}^{H}\mathbf{\Phi} \mathbf{H}_{br})\mathbf{F}||_2^2 \\
    \text{s.t.} & \quad ||\mathbf{F}||^2 = P_{max} \\
               & \quad |\mathbf{\Phi}_{i,i}|^2 = 1.
\end{align*}

For this MISO system, we know that, for a fixed $\mathbf{\Phi}$, the optimal beamforming vector that maximizes the SNR is given by the maximum-ratio transmission (MRT) method \cite{wu2019MRT} as follows:
\begin{align}
    \mathbf{F}^* = \frac{\sqrt{P_{max}} \mathbf{H}}{||\mathbf{H}||_2}.
\end{align}

Hence, the optimization problem can be simplified as:
\begin{align}
    (P_1) : \max_{\mathbf{\Phi}} & \quad ||\mathbf{H}_{bu}^{H} + \mathbf{H}_{ru}^{H}\mathbf{\Phi} \mathbf{H}_{br}||_2^2 \\
    \text{s.t.} & \quad |\mathbf{\Phi}_{i,i}|^2 = 1, \quad i=1,\dots,M.
\end{align}
The optimization problem as formulated in ($P_1$) is a non-convex problem due to the unimodular constraints.

Several works attempted to solve this optimization problem. In particular, \cite{rehman2021VAMP} focused on minimizing the Minimum Mean Square Error (MMSE) criterion which is equivalent to maximizing the user SNR according to the I-MMSE relationship. The authors derived a closed-form solution for the beamforming vector and proposed a modified algorithm based on VAMP algorithm. This work shows better performance compared to ADMM and has a complexity of $O(NM + M^2 + N^3)$. The disadvantages of this method are two-fold: (i) it is an iterative algorithm, and (ii) it requires the perfect knowledge of CSI. In \cite{yu2019SDR}, the authors used an SDR method which is characterized by a high complexity (i.e. $O((M + 1)^6)$).

To overcome the above-mentioned shortcomings of traditional optimization methods, DRL is used to learn the optimal phase shifts. The advantage of DRL compared to the previous methods is, it can be queried in real-time and necessitates less CPU time. A comparison between SDR and DRL-based solutions for ($P_1$) was presented in \cite{feng2020deep}. In this work, we provide a comparison between VAMP, ADMM, and DRL in addition to a performance analysis under different channel impairments.

\section{DRL-Based Approach}

\subsection{Background}
The standard reinforcement learning framework consists of an agent learning to maximize its expected cumulative rewards through interacting with an environment. At each interaction or timestep $t$, given an observation $s_t$, the agent chooses an action $a_t$ and receives an instantaneous reward $r_t$, and the system transit to a new state $s_{t+1}$. The tuples $(s_t, a_t, r_t, s_{t+1})$ constitute the agent experiences used for learning the optimal policy $\pi : S \mapsto \mathcal{P}(A)$. This learning problem is often modeled as a Markov Decision Process (MDP), described by the tuple $(S, A, P, R, \gamma)$, when the state space is fully-observable. $S$ and $A$ define the state and the action spaces,
respectively; $P := S \times A \mapsto [0, 1]$ denotes the transition probability function, $R := S \times A \times S \mapsto R$ is the reward function, and $\gamma \in [0, 1]$ is a discount factor that trades-off the immediate and upcoming rewards.

We define the agent's expected return as the sum of the \emph{discounted} future rewards $R_t= \mathbb{E} \Big[\sum_{i=t}^{T} \gamma^{(i-t)} R(s_i, a_i, s_{i+1})|a_i \sim \pi(.|s_i) \big] $. Thus, the agent's objective is to maximize the expected long-term reward $J = \mathbb{E}_{\pi}\Big[\sum_{t=0}^{\infty} \gamma^t R_t\Big].$

Another well-known function used to measure the agent's returns is the action-value function $Q$. It measures the expected accumulated rewards after executing an action $a_t$ at a state $s_t$ and following the policy $\pi$ thereafter:
\begin{equation*}
\resizebox{.92\hsize}{!}{$Q^{\pi}(s,a) = \mathbb{E} \Bigg[\sum_{t=0}^{\infty} \gamma^t R(s_t, a_t, s_{t+1}) | a_t \sim \pi(.|s_t), s_0=s, a_0=a \Bigg]$}.
\end{equation*}

The famous $Q-$learning algorithm makes use of the \emph{Temporal Difference} (TD) formulation where the $Q$ function is learned using the following recursive expression
\begin{equation}
\label{eq:tdupdate}
\resizebox{.91\hsize}{!}{$Q(s_t,a_t) = (1-\alpha)Q(s_t,a_t) + \alpha \big[r(s_t,a_t) + \gamma \max_{a^\prime} Q(s_{t+1},a^\prime)\big]$}
\end{equation}
in which $\alpha$ is a learning rate.

The main disadvantage of the $Q$-learning method is the $\max$ operator in (\ref{eq:tdupdate}) which limits it to discrete action spaces. In this context, Deterministic Policy Gradient (DPG) algorithm \cite{silver2014DPG} can be considered as an extension of $Q$-learning to continuous action spaces by replacing the $\max_{a^{\prime}} Q(s_{t+1}, a^{\prime})$ in (\ref{eq:tdupdate}) by $Q(s_{t+1}, \mu(s_{t+1}))$, where $\mu(s_{t+1}) = \argmax_{a^{\prime}} Q(s_{t+1}, a^{\prime})$. Thus, DPG algorithms concurrently learn a $Q$-function $Q_{\phi}$ and a policy $\mu_{\psi}$, where $\phi$ and $\psi$ are the parameters of the $Q$-function and the policy, respectively. The $Q$-function is learned by minimizing the \emph{Bellman error} in (\ref{eq:5}). As for the policy, the objective is to learn a \emph{deterministic policy} that outputs the action corresponding to $\max_a Q_\phi(s,a)$. The policy is called deterministic because it gives the exact action to take at each state $s$. Hence, the learning process consists in performing gradient ascent with respect to $\theta$ to solve the objective in (\ref{eq:6}):

\begin{align} 
    \phi^* &=  \argmin_\phi \frac{1}{2} \sum_{(s,a,r,s^{\prime})} ||Q_{\phi}(s, a) - y||^2 \label{eq:5} \\ 
    y &=  R(s,a) + \gamma \hat{Q}_{\phi^\prime}(s^{\prime},\hat{\mu_{\psi^{\prime}}}(s^{\prime})) \label{eq:dqntarget} \\
    J(\theta) &= \mathbb{E}_{s \sim D} \bigg[ Q_\phi(s,\mu_\theta(s)) \bigg] \label{eq:6} \\
    \nabla_\theta J(\theta) &= \mathbb{E}_{s \sim D} \bigg[\nabla_{\theta} \mu_{\theta}(s) \nabla_{a} Q_\phi(s,a)|_{a=\mu_{\theta}(s)}\bigg] \label{eq:actorUpdate}
\end{align} 
where $\hat{Q}$ and $\hat{\mu}$ are called target networks which are copies of the $Q$ and $\mu$.These target networks are used to compute the target values [\ref{eq:dqntarget}]. The target parameters are updated using a Polyak update with a parameter $\tau$.

Deep Deterministic Policy Gradient (DDPG) \cite{lillicrap2015DDPG} has been widely used to solve wireless communication problems. It is an extension of the DPG method where the policy $\mu_\psi$ and the critic $Q_{\phi}$ are both deep neural networks.

\subsection{MDP Formulation}
The optimization problem in ($P_1$) is modeled as an MDP where the agent is the BS. In what follows, we define the state and action spaces and the reward function:
\begin{itemize}
    \item \textbf{State space $S$}: consists of the user SNR and action $s_t = [\rho_{t-1}, a_{t-1}]$. The dimension of state space  is $M+1$ and spans $\mathbb{R}_{+}$. Using the past actions helps mimicking the recurrent version of RL.
    \item \textbf{Action space $A$}: at each timestep, the agent chooses the phase shifts $\theta_i, i=1\dots M$. Thus, the dimension of the action space is $M$ and $a_t = [\theta_1, \dots, \theta_M] \in [0, 2\pi]^M$.
    \item \textbf{Reward}: at each time step, the agent's reward is the received SNR as defined in (\ref{eq:snr}).
\end{itemize}
Technically, this formulation can be considered as a partially observable MDP since the agent observes only its SNR and does not have access to the CSI which is used to compute the next state. So, the agent needs to infer the CSI in order to recover the full state. In this paper, we will solve this problem by using a model-free RL algorithm. In future works, other methods can be considered to deal with the  problem of partial observability of the environment. 

We assume that the agent interacts with the environment in an episodic manner. An episode ends when the number of time-steps exceeds a fixed horizon $T$. At the beginning of each episode, we sample new channel realizations and randomly choose the phase shifts to compute the first state $s_0$. 

The  pseudo-code in \textbf{Algorithm 1} summarizes the DDPG algorithm to find the optimal phase shifts. To ensure a better exploration, we add an adaptive parameter space noise \cite{plappert2017parameterNoise} in addition to the action space noise. The parameter noise is applied to the weights of the actor network. The traditional action noise makes the policy stochastic and enables the perturbation of action likelihoods independently from the state. However, the parameter space noise incorporates the noise directly in the parameters yielding a \emph{state-dependent} exploration \cite{ruckstiess2008SDE}. For more details, we refer the reader to \cite{plappert2017parameterNoise}.

\begin{algorithm}
\small
\caption{DDPG with Parameter Noise Exploration}\label{algo:DDPG}
\begin{algorithmic}[1]
\State \textbf{Input}: Initialize actor parameters $\psi$, critic parameters $\phi$, empty memory buffer $\mathcal{D}$, parameter space noise standard deviation $\sigma_0$,  intervals for training and noise adaptive scaling,  $T_{\text{train}}$, $T_{\text{adapt}}$, number of training steps $N_{\text{train}}$, an adaption update coefficient $\alpha$ and threshold $\delta$
\State Initialize the target networks parameters  $\phi^{\prime} \leftarrow \phi$ and $\psi^{\prime} \leftarrow \psi$\label{algo:init1}
\State Set $\tilde{\mu} = \mu_{\tilde{\psi}}$ for exploration and $\bar{\mu} = \mu_{\bar{\psi}}$ for noise adaption, $t=0$
\Repeat
\State Apply noise perturbation to actor parameters $\tilde{\psi} \leftarrow \psi + \mathcal{N}(0, \sigma_k^2\mathbf{I})$ to obtain $\tilde{\mu}$
\State Receive state $\bm{s_t}$ and choose an action $\bm{a_t} = \text{clip}(\tilde{\mu}(\bm{s_t})+\epsilon, A_{\text{low}}, A_{\text{high}})$ according to the perturbed policy $\tilde{\mu}$ where $\epsilon \sim \mathcal{N}(0.\sigma_a)$
\State Execute $\bm{a_t}$ and observe the receive reward $r$, the next state $\bm{s^{t+1}}$. Store $(\bm{s_t}, \bm{a_t}, r, \bm{s_{t+1}})$ in $\mathcal{D}$
\State If $\bm{s^{t+1}}$ is terminal, reset the environment and update the perturbed actor $\tilde{\psi} \leftarrow \psi + \mathcal{N}(0, \sigma_k^2\mathbf{I})$
\State $t=t+1$
\If {$t$ mod $T_{\text{train}}$}
\For {$i=0,\dots, N_\text{train}$}
\State Sample a minibatch $\mathcal{B}$ from $\mathcal{D}$
\State Update the actor and the critic parameters as specified by (\ref{eq:5}), (\ref{eq:actorUpdate})
\State Update target networks
\State  \quad \quad ${\phi^{\prime}} \leftarrow \tau \phi+(1-\tau) \phi^{\prime}$ \label{algo:update-critic-target}
\State \quad \quad $\psi^{\prime} \leftarrow \tau \psi+(1-\tau) \psi^{\prime}$ \label{algo:update-actor-target}
\If{$i$ mod $T_{\text{adapt}}$}
\State Perturb $\bar{\psi} \leftarrow \psi + \mathcal{N}(0, \sigma^2_k\mathbf{I})$ and obtain $\bar{\mu}$
\State Estimate the standard deviation between $\mu$ and $\bar{\mu}$
\State \quad $d_k = \mathbb{E}_{\bm{s}}\big[ \sqrt{\frac{1}{|\mathcal{A}|}\sum_{j=}^{|\mathcal{A}|}(\mu(\bm{s})_j-\bar{\mu}(\bm{s})_j)^2}\big]$
\State Adapt the parameter noise scale
\State \quad $\sigma_{k+1}\leftarrow 
\left\{\begin{matrix}
 \alpha\,\sigma_k \quad \text{ if } d_k \leq \delta, \\ 
\frac{1}{\alpha}\sigma_k  \quad \text{otherwise}
\end{matrix}\right.$
\EndIf
\EndFor
\EndIf
\Until convergence 
\end{algorithmic}
\end{algorithm}

\section{Numerical Results}
\subsection{Simulation Setup}
In this section, we detail the experimental setup used
to generate the results.

\noindent\textbf{Channel models}:
\noindent The channel between the BS and the user is assumed to experience Rayleigh fading which suggests that the line-of-sight (LOS) signal between them is blocked:
\begin{align*}
    \bm{H}_{bu} = \sqrt{G_{bu}} \bm{h}_{bu}
\end{align*}
where $\bm{h}_{bu} \in \mathbb{C}^{N\times1}$ and its elements are sampled independently from $\mathcal{CN}(0,1)$. The channel between the BS and IRS as well as the one between the IRS and the user are Ricean fading:
\begin{align*}
    \bm{H}_{br} &=\sqrt{G_{br}} \Big(\sqrt{\frac{K_1}{K_1+1}}\bm{H}_{br}^{LOS}+\sqrt{\frac{1}{K_1+1}}\bm{H}_{br}^{NLOS} \Big)  \\
    \bm{H}_{ru} &=\sqrt{G_{ru}} \Big(\sqrt{\frac{K_2}{K_2+1}}\bm{H}_{ru}^{LOS}+\sqrt{\frac{1}{K_2+1}}\bm{H}_{ru}^{NLOS} \Big)
\end{align*}
where $\bm{H}_{br}^{LOS}$ and $\bm{H}_{ru}^{LOS}$ are the deterministic LOS components for IRS-user and BS-IRS links computed as shown \cite{feng2020deep}. $\bm{H}_{br}^{NLOS}$ and $\bm{H}_{ru}^{NLOS}$ are the NLOS components independently distributed as $\mathcal{CN}(0,1)$. $K_1$ and $K_2$ represent the Ricean factors. 

    


\noindent The path-loss model is given by $G(d) = G_0 - 10 \alpha \log_{10}(d/d_0)$ in dB, where $G_0$ is the 
path-loss at the reference distance, $d_0$ is the reference distance and $\alpha$ is the path loss exponent.

\noindent\textbf{Simulation hyperparameters}:
\noindent Table \ref{tab:params} summarizes the simulation parameters. For the DDPG algorithm, we use the hyperparameters reported in \cite{feng2020deep} in addition to the parameter space noise. We find that using a long episode horizon (i.e $1000$) is critical for learning. 

\noindent\textbf{Policy representation}:
For both actor and critic, the same architecture is used: a feedforward deep neural network with two hidden layers with $300$ and $200$ hidden units and Relu activations. Before applying the activation function, we use a normalization layer \cite{ba2016layerNorm} to stabilize the learning under parameter noise. 

\noindent\textbf{Baselines}:
We compare the DRL agent with VAMP and ADMM algorithms due to their convergence guarantees and speed. 

For reproducibility purposes, we will release the code of the environment used for the simulations below. \cite{stable-baselines} is used as an implementation of the DRL algorithm.

\begin{table}[]
\caption{Simulation parameters}
    \label{tab:params}
    \centering
    \begin{tabular}{c|c}
    \hline 
    Parameter & Value \\
    \hline
         $K_1= K_2$& $10$ \\
         $N$& $10$ \\
         $M$& $50$ \\
         $\sigma^2_n$ & $-80$ dBm \\
         $P_{\max}$ & $5$ dBm \\
         $d_0$ & $1$ m \\
         $G_0$ & $-30$ dB \\
         $\alpha_{br}, \alpha_{bu}, \alpha_{ru}$ & $2,2, 2.8$\\
         $d_{br}, d_{bu}, d_{ru}$ & $51, 48, 1.5$ m\\
         \hline
         Hidden layers & $300,200$\\
        learning rate & $1e^{-3}$ \\
        Episode horizon $H$ & $1000$ \\
        Replay buffer size & $1e^6$ \\
        $N_{train} = T_{adapt}$ & $50$ \\
        $T_{train}$ & $1000$ \\
        $\sigma_a = \sigma_0$ & $0.1$\\
        \hline
    \end{tabular}
\end{table}

\subsection{Results and Discussion}
\subsubsection*{\textbf{Performance comparison}}
\begin{figure}[ht!]
    \centering
    \includegraphics[scale=0.5]{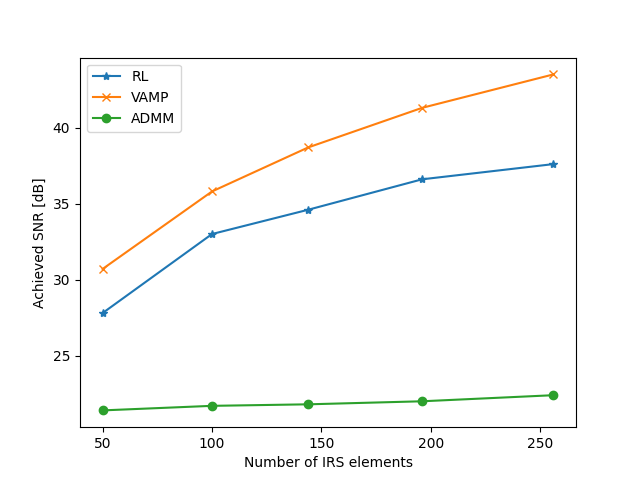}
    \caption{Achieved SNR.}
    \label{fig:snr}
\end{figure}

\begin{figure}[ht!]
    \centering
    \includegraphics[scale=0.5]{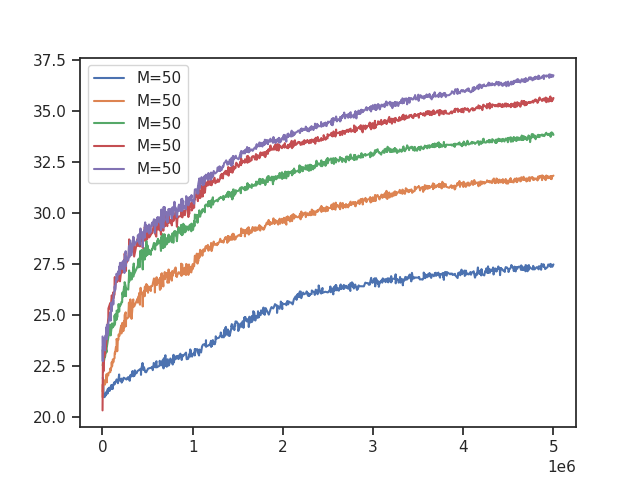}
    \caption{DRL training curves for different number of IRS elements.}
    \label{fig:train}
\end{figure}

We first compare the three considered methods: VAMP, ADMM, and DRL. Here, we set the parameters of the network as reported in Table \ref{tab:params} and vary the number of IRS elements. In Fig.~\ref{fig:snr}, we illustrate the achievable SNR as a function of the number of IRS reflective elements for the different considered algorithms. The achievable SNR increases as the number of IRS elements increases. With perfect CSI, VAMP has the best achieved SNR compared to the other methods. We observe that the DRL method is able to achieve good performance compared to ADMM even without access to channel information. Fig.~\ref{fig:train} shows that the agent's reward improves as the number of IRS elements increases. However, the SNR gain becomes smaller as the number of IRS elements gets higher.

Furthermore, we examine the inference speed of each algorithm and summarize them in Table \ref{tab:inference_time}. We average the inference time over $1000$ realizations. VAMP and ADMM algorithms are run for $30$ iterations to achieve convergence \cite{rehman2021VAMP}. After training, the DRL agent is queried in real-time to get the optimal phase shifts. Compared to traditional optimization techniques, DRL is superior in terms of inference speed which makes it more appealing for practical implementations.

\begin{table}[ht!]
 \caption{Inference time in ms}
    \label{tab:inference_time}
    \centering
    \begin{tabular}{c|c|c|c}
\hline
 M & VAMP &ADMM&DRL\\
 \hline
 50  & 30 & 19 & 0.3 \\
 100 & 38 & 29 & 0.48 \\
 144 & 48 & 41 & 0.49 \\
 196 & 76 & 64 & 0.6 \\
 256 & 100 & 100 & 0.62 \\
 \hline
\end{tabular}
\end{table}

\subsubsection*{\textbf{Performance under unreliable channel}}
Since in real-world scenarios, the CSI acquired by the transmitter is not perfect and subject to errors, in this section we investigate the impact of CSI errors on the performance of the considered algorithms. 
As the first step, we obtain the optimal phase shifts and beamforming vector using the available CSI. Afterwards, we apply perturbations to the channels as shown in (\ref{eq:perturbation}):
\begin{align}\label{eq:perturbation}
\bm{H}[t] = \sqrt{\epsilon}\bm{H}[t-1] + \sqrt{\epsilon-1}\bm{w}[t] 
\end{align}
where $\bm{w}[t]$ contains independent and identically 
distributed $\mathcal{CN}(0,1)$ elements. We compute the user SNR using the optimal phase shifts and the perturbed channels. Note that the objective is to study the impact of the channel errors on the optimal solutions. This means that we verify if the obtained solutions remain optimal with small changes in the CSI.

\begin{figure}[ht!]
    \centering
    \includegraphics[scale=0.53]{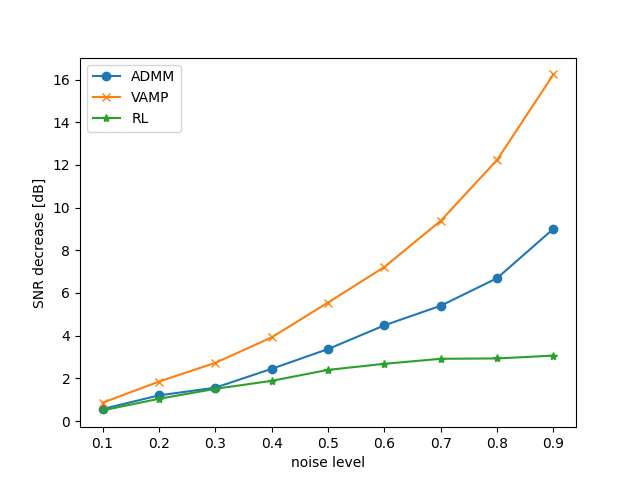}
    \caption{The loss in SNR due to channel errors.}
    \label{fig:channel_noise}
\end{figure}

Fig.~\ref{fig:channel_noise} illustrates the decrease in the user SNR due to different levels of noise. Since VAMP and ADMM rely on perfect CSI to find optimal solutions, the drop in performance is severe compared to DRL. Indeed, the DRL agent is able to learn a good policy without relying on the CSI directly which makes it more robust to changes in the environment.  Furthermore, we observe that VAMP is more sensitive to changes in the channel than ADMM. For a high level of noise, VAMP performance drops by more than $16$ dB whereas the performance of the DRL agent decreases only by $3$ dB.

\subsubsection*{\textbf{Performance under user mobility}}
\begin{figure}[t!]
    \centering
    \includegraphics[scale=0.5]{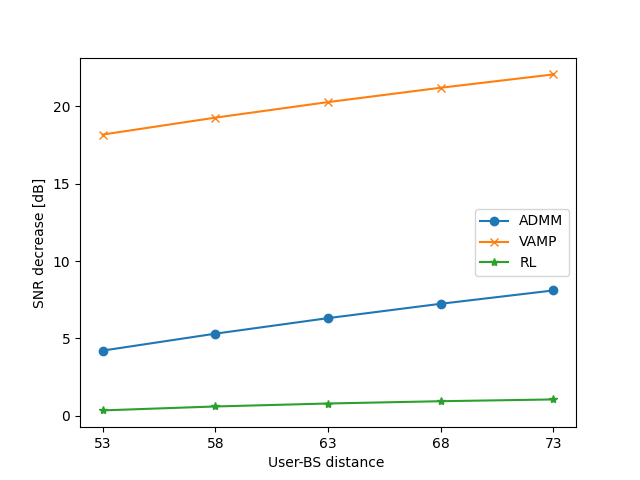}
    \caption{The loss in SNR due to user mobility.}
    \label{fig:pl}
\end{figure}

To further verify the robustness of the DRL to changes in its environment, we increase the distance between the BS and the user to simulate a mobile target. Originally, the distance between the user and BS is $48$ m. We assume that the user is moving away from the BS which represents the worst-case scenario. Hence, the distance is increased by a step-size in $[5,25]$m. Note that we first compute the optimal phase shifts and beamforming vector using the current CSI, then we increase the user distance and compute the SNR using the optimal solutions. 
A similar pattern is observed in Fig. \ref{fig:pl} where VAMP and ADMM suffer dramatic drops in the user SNR. Surprisingly, the DRL performance stays approximately the same.


\section{Conclusion}
In this paper, we aimed to investigate the robustness of DRL against changes in CSI. This is a practical scenario where the CSI at the transmitter is subject to errors and noise. DRL shows an interesting resilience to different channel impairments. This is advantageous compared to other optimization/inference methods where perfect CSI is crucial to ensure good performance. Furthermore, we provide a rigorous comparison between DRL, VAMP, and ADMM to optimize the phase shift matrix for downlink MISO transmission. Numerical results show that DRL is better than ADMM which fails to find a good solution for the considered system. Although VAMP has better performance, the DRL agent is capable of learning good policies without relying on any CSI. 

In this work, we have only considered errors in the CSI; however, other sources of perturbations can impact the performance of the DRL agent. In a future work, we will do a more detailed investigation of the DRL methods for wireless continuous control. Furthermore, only robustness during inference is considered in this paper. Training under uncertainty can also be studied as an extension.
\bibliographystyle{ieeetr}
\bibliography{refs}

\end{document}